%% file: ms.tex
\documentclass{article}
\pdfoutput=1
\usepackage{arxiv}
\usepackage[authoryear]{natbib}
\usepackage[utf8]{inputenc} 
\usepackage[T1]{fontenc}    
\usepackage{hyperref}       
\usepackage{url}            
\usepackage{booktabs}       
\usepackage{amsfonts}       
\usepackage{nicefrac}       
\usepackage{microtype}      
\usepackage{lipsum}		
\usepackage{graphicx}
\usepackage{natbib}
\usepackage{doi}

\usepackage{algorithm}
\usepackage{algpseudocode}
\usepackage{makecell}  
\usepackage{comment}

\title{On the Nature of the Phenotype in Tree Genetic Programming}

\author{ \href{https://orcid.org/0000-0002-6382-3245}{\includegraphics[scale=0.06]{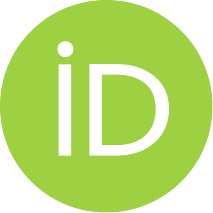}\hspace{1mm}Wolfgang Banzhaf} \\
	Department of Computer Science and Engineering\\
	Michigan State University\\
	East Lansing, MI 48824 \\
	\texttt{banzhafw@msu.edu} \\
	\And
	\href{https://orcid.org/0000-0002-6458-942X}{\includegraphics[scale=0.06]{orcid.pdf}\hspace{1mm}Illya Bakurov} \\
	Department of Computer Science and Engineering\\
	Michigan State University\\
	East Lansing, MI 48824 \\
	\texttt{bakurov1@msu.edu} 
}

\begin{document}
\maketitle

\begin{abstract}

In this contribution, we discuss the basic concepts of genotypes and phenotypes in tree-based GP (TGP), and then analyze their behavior using five benchmark datasets. We show that TGP exhibits the same behavior that we can observe in other GP representations: At the genotypic level trees show frequently unchecked growth with seemingly ineffective code, but on the phenotypic level, much smaller trees can be observed. To generate phenotypes, we provide a unique technique for removing semantically ineffective code from GP trees. The approach extracts considerably simpler phenotypes while not being limited to local operations in the genotype. We generalize this transformation based on a problem-independent parameter that enables a further simplification of the exact phenotype by coarse-graining to produce approximate phenotypes. The concept of these phenotypes (exact and approximate) allows us to clarify what evolved solutions truly predict, making GP models considered at the phenotypic level much better interpretable.
\end{abstract}




\maketitle

\section{Introduction}
\label{s_intro}

\input{intro}

\section{Background}
\label{s_background}

\input{background}

\section{The Phenotype in Tree Genetic Programming}
\label{s_approach}

\input{approach}

\section{Experiments}
\label{s_experiments}

\input{experiments}

\section{Conclusions}
\label{s_conclusion}

\input{conclusions}


\newpage

\bibliographystyle{plainnat}
\bibliography{refs}

\end{document}

%% file: intro.tex
Since early in its development, GP has 
been associated with Machine Learning tasks \citep{b92_gp_koza,b98_gp_intro_banzhaf}. 
What is special about GP compared to other EC approaches is that its expected result is an active solution (the program, algorithm, formula, model, etc.) that itself behaves in some way to process inputs into outputs. Thus it is not enough to simply consider a static solution to an optimization problem, but what is required is the ability to modify the solution in both its content and its complexity. The genetic operators for GP must therefore be able to grow and shrink the complexity of solutions in the evolutionary process, an ability that comes with its particular challenges. 

All this is well known and has led to some surprises early on in the history of GP \citep{b92_gp_koza, b94_gp_intelligence_angeline, b98_gp_intro_banzhaf}: To the bafflement of many researchers, a substantial portion of the code produced by GP was ineffective (neutral, non-effective, intron code, junk, etc.), which did not seem to play a role in the actual behavior of the evolved algorithms, but nevertheless was produced by the evolutionary process. Researchers found that if they waited long enough in evolutionary runs, the overwhelming majority of the code was of this type, a phenomenon coined 'bloat'.

Several theories have been put forward as to what the reasons are for this apparent waste of code material with its attendant strain on computational resources of the evolutionary process (both memory and compute cycles). 
A very interesting consequence of ineffective code in GP is that it necessitates the discernment of genetic material (genotypes) and behavioral determinants (phenotypes). Allowing for some of the genetic material to not having an impact on behavior requires the introduction of this difference and a mapping between genotypes and phenotypes. In contrast to popular belief, most GP representations today have such a mapping, even if it is somewhat hidden from view. 

Three of the most typical representations used in GP are expression trees in tree GP (TGP), linear sequences of instructions in linear GP (LGP) and circuit-type graphs in cartesian GP (CGP). Here we shall be mostly concerned with TGP, and want to only briefly touch the other two representations. All three require genotypes (expression trees in TGP, numbers encoding instructions in LGP, numbers encoding graphs in CGP) and phenotypes (expression trees in TGP, instruction sequences in LGP and graphs in CGP). To draw a clear distinction: genotypes are manipulated by genetic operations while phenotype are the encoded structures that provide the behavioral function of the algorithms.
Because the genotypes are normally larger than the phenotypes, taking into account that some of the code is non-effective or neutral (either structurally or semantically), it is the phenotypes that can provide an understanding of what the algorithms do. 

In a Machine Learning (ML) context, GP provides a user with a completely different experience in terms of transparency and interpretability of the final models when compared to "black-box" models of, say, deep neural networks. By manipulating discrete structures, the inner workings of the model are clearer, and the relationships between features and predictions are explicitly defined in GP. As a result, evolved solutions can offer insights into how the model arrives at its predictions, allowing humans to understand the decision-making process. 
Although GP is not the only "white-box" method in the spectrum of ML methods, it allows for an unprecedented degree of flexibility of representations and modularity, making it suitable for numerous different ML tasks: classification~\citep{c14_m3gp_ignalalli,c15_m3gp_munoz,a19_m4gp_lacava}, regression~\citep{c12_gsgp_moraglio,a19_gsgp_framework_castelli,c16_epsilon_lexicase_lacava}, feature engineering~\citep{a12_feature_construction_neshatian,a19_feature_construction_tran}, manifold learning~\citep{a22_manifold_lensen}, active learning~\citep{c23_active_learning_gp_hat}, image classification~\citep{p94_recombination_selection_construction_tackett,c04_multi_class_image_zhang,c17_gp_skin_cancer_ain,a17_gp_image_class_iqbal}, image segmentation~\citep{a23_ssnet_stack_gp_bakurov}, image enhancement~\citep{a22_elaine_correia,c21_gp_evolving_image_enhancement_correia}, automatic generation of ML pipelines~\citep{a19_tpot_gp_olson} and even neural architecture search~\citep{c15_sml_goncalves,a19_denser_assuncao}. 
The area of explainable AI is consistently receiving more attention from both practitioners and researchers~\citep{a21_role_exai_markus,a23_exai_ethics_pekka} and GP has gained popularity where human-interpretable solutions are paramount. Real-world examples include medical image segmentation~\citep{a23_kartezio_med_image_segment_cortacero}, prediction of human oral bioavailability of drugs~\citep{a12_bloat_free_gp_bioav_silva}, skin cancer classification from lesion images~\citep{a22_gp_skin_cancer_qurrat}, and even conception of models of visual perception~\citep{a23_friqa_gp_bakurov}, etc. 
Besides being able to provide interpretable models, there is evidence that GP can also help to unlock the behaviour of black-box models~\citep{c19_gp_insite_blackbox_evans,a23_ssnet_stack_gp_bakurov}. 

Given that GP phenotypes are even smaller than genotypes, interpretability is enhanced by focussing on phenotypes when analysing GP models.
In this contribution, we therefore study the relationship between genotypes and phenotypes in TGP, and develop a unique simplification technique that allows us to remove semantically ineffective code from GP genotype trees based on numerical simplification. 
Particularly, in this work, we show that:
\begin{itemize}
    \item Trees of TGP represent both the genotype and phenotype of individuals.
    \item Phenotypes are hidden within the larger genotypic trees but can be extracted via simplification without affecting their behavior, which facilitates understanding of the models.
    \item Phenotypes can be further simplified to shrink model size but this might impact generalization ability.
    \item We observe the population dynamics of genotypes and phenotypes throughout the evolutionary process to facilitate our understanding of the genotype-phenotype relationship.
\end{itemize}

The paper is organized as follows: Section~\ref{s_background} introduces the necessary theoretical background by providing a brief discussion of the bloat phenomenon in TGP, an analogy to phenomena in living systems, some approaches to remedy this situation and a generic look at the difference between genotypes and phenotypes. Section~\ref{s_approach} describes our algorithm to extract phenotypes from genotypes in TGP. We also introduce the notion of an approximate phenotype. Section~\ref{s_experiments} characterizes the datasets used in our study, discusses the hyper-parameters used, and shows the results obtained. 
Finally, Section~\ref{s_conclusion} draws the main conclusions and proposes future research ideas.

%% file: background.tex

As a natural result of genetic operators applied to tree genotypes, GP trees tend to grow in size during evolution in search for a better match to the desired behaviour. 
However, this growth is not always justified by an improvement in performance (such as the generalization ability on unseen data). It has long been observed that standard TGP tends to produce excessively large and redundant trees without a corresponding improvement in terms of fitness~\citep{b92_gp_koza,b94_gp_intelligence_angeline}. Moreover, it was shown that beyond a certain program length, the fitness distribution of individuals converges to a limit~\citep{b02_gp_foundations_langdon}. 
The apparent unnecessary growth of GP trees is often called the bloat phenomenon and several justifications were proposed for its existence. 
For decades, researchers tried to address this problem. We briefly review part of this voluminous literature here, but cannot address it in its full breadth. 

The consequences of bloat in TGP are hard to underestimate: the amount of computer memory to store a population grows and concomitantly the evaluation time of trees increases, making the system less attractive for real-time applications. Moreover,  unnecessarily large trees (genotypes) make visual inspection and interpretation difficult, if not impossible. In LGP and CGP the mechanisms for bloat result mainly in structural ineffective code. The difficulty for TGP is that its bloat is mostly semantic \citep{banzhaf2023}.

\subsection{Bloat}
\label{ss_back_bloat}
In binary-string Genetic Algorithms (GAs), it was found that highly fit building blocks become attached, by coincidence, to adjacent unfit building blocks. As a result, these joint entities were propagated throughout the population using recombination, subsequently preventing highly fit building blocks at adjacent locations from joining~\citep{c93_hitchhiking_mitchell}. W. Tackett demonstrated that this phenomenon, called 'hitchhiking', also exists in TGP and that it was a consequence of recombination acting in concert with fitness-based selection~\citep{p94_recombination_selection_construction_tackett}.

In~\citep{c98_fitness_causes_bloat_langdon, c98_fitness_causes_bloat_mutation_langdon}, W. Langdon and R. Poli show that fitness-based selection makes the evolutionary search converge to mainly finding candidate solutions with the same fitness as previously found solutions. In the absence of improved solutions, the search may become a random search for new representations of the best-so-far solution. Given the fact that the variable-length representation of TGP allows many more representations of a given behaviour to be longer, these solutions are expected to occur more often and representation length naturally tends to increase. In other words, a straightforward fitness-based approach contributes to bloat. These findings are in agreement with the earlier findings of W. Tacket~\citep{p94_recombination_selection_construction_tackett} who concluded that (i) the average growth in size is proportional to the underlying selection pressure, and that (ii) bloat does not occur when the fitness is completely ignored.

In~\citep{c07_crossover_bias_theory_dignum} the authors provide a theoretical model of program length distribution in the population assuming a repeated application of subtree crossover (i.e., crossover with a uniform selection of points) on a flat fitness landscape. The model is validated empirically and it is proved that the reproduction process has a strong bias to sample shorter programs when crossover with a uniform selection of points is applied. It was demonstrated that, if a reasonable minimum program size exists for relatively fit programs, the repeated application of fitness-based selection and subtree crossover will cause bloat to occur and that this phenomenon will be more acute if allowing crossover to distribute a population before applying fitness-based selection.

A number of researchers demonstrated that bloat is an evolutionary response to protect solutions from the destructive effects of operators, with crossover the main focus of their studies~\citep{c95_accurate_replication_in_gp_mcphee,c94_gp&redundancy_blickle,b94_evolution_of_evolvability_altenberg,c95_complexity_compression_nordin,a02_causes_of_growth_soule}.
In particular, when a program contains large sections of code that do not have a significant effect on its behaviour (introns), the application of a given genetic operator in that section is expected to have no overall effect on the program's behaviour (i.e., it is neutral). Several studies demonstrate that genetic operators are more likely to deteriorate offspring fitness than increase it and that a large proportion of recombinations are fitness-neutral~\citep{b96_explicitly_defined_introns_nordin,b99_evolution_of_size&shape_langdon,a02_causes_of_growth_soule}.
Thus, by increasing the proportion of introns to effective sections of code (exons), evolution increases the likelihood that the product of genetic operators will be neutral in terms of fitness \citep{banzhaf2002some}.
However, if bloat makes it more difficult to deteriorate the fitness of a solution, it also makes it harder to improve it. In this sense, bloat constitutes a serious problem for a sustainable evolutionary process. 



\subsection{A Biological Analogy}
\label{ss_back_bio}

EAs are strongly inspired by evolutionary biology and it is not surprising that the topic addressed in this manuscript can be also found there. 
The analogue for the "redundancy" or "bloat" in genetics is called non-coding DNA and it refers to all DNA that is not involved in the coding for amino acids 
(i.e., does not provide instructions for making proteins). It is considered that the non-coding DNA makes up 97-99\% of  total DNA~\citep{c96_introns_genetics_wu,b22_human_biochemistry_litwack}. 
Initially, biologists thought that it had no function and referred to the non-coding DNA as "junk". 
Later, it was found that many of the non-coding sequences are repeated movable elements that facilitate genomic rearrangements; also, it was found to contain sequences that act as regulatory elements, determining when and where genes are turned on and off~\citep{a17_non_coding_rna_not_junk}. Finally, it was found that alternative splicing of sequences at exon/intron boundaries ~\citep{a78_exon_theory_gilbert} can provide for a combinatorial variety of proteins derived from the same gene sequence. 

All of this suggests that non-coding DNA is of evolutionary importance. Otherwise, one would expect it to have been eliminated by natural selection long ago due to the extra energy it requires for sustaining and processing~\citep{c96_introns_genetics_wu}. In turn, one could suspect that the non-coding regions of GP genotypes might also serve a useful role in artificial evolution, which gives support to the neutrality for protection hypothesis.

\subsection{Simplification}
\label{ss_back_counteract}
Here we consider the concept of the phenotype in TGP as that part of an individuum that produces overall behavior. We use a simplification algorithm to extract the phenotype tree from a genotype tree, and keep them side by side so as to be able to monitor their development over the course of evolution.
Simplification algorithms have been traditionally used in TGP, but not with a concept of phenotype in mind.
We divide such contributions into two categories: (i) those that have no effect on the evolutionary trajectory of a run (observing approaches); and (ii) those that do affect the evolutionary trajectory (intervening approaches).
As pointed out in \citep{a22_simplification_gp_javed}, historically, the former approaches were first, but quickly overtaken by the latter. Javed et al. call the former 'offline' approaches vs. the latter called 'online'.

Under category (i), we count ''offline simplification'', properly defined as "the analysis and reduction in complexity" of an individual outside the evolutionary run, e.g. at the end of a run~\citep{b92_gp_koza,a10_dormancy_gp_jackson,c14_effective_simplification_spector,a20_pruning_gp_test_rockett}, or during a run, but not given back to evolution - the latter we call ''monitoring''~\citep{c95_complexity_compression_nordin,c08_algebraic_simplification_gp_mori}. What researchers have done here, without saying it, is to observe the phenotype of an individual similar to what we do here, without explicitly calling it that.

Under category (ii), we count ''online simplification'' or ''pruning''~\citep{c06_algebraic_bloat_gp_wong,a20_semantic_approximation_gp_nguyen,a09_simplification_bb_gp_kinzett,a10_bloat_control_diversity_gp_alfaro,a10_dormancy_gp_jackson}, as well as limiting the size of individuals in various ways~\citep{b92_gp_koza, b99_evolution_of_size&shape_langdon, c07_crossover_bias_theory_dignum,c08_crossover_bloat_limit_dignum,a09_dynamic_limits_bloat_silva}, operator equalization and its variants~\citep{c08_operator_equalization_dignum, c11_operator_equalization-r_silva, c09_operator_equalization-m_silva}, time-based
approaches to controlling bloat~\citep{c20_time_bloat_gp_vega,a20_time_duration_gp_vega}, fitness and selection approaches (like double tournaments etc.)~\citep{c03_simple_bloat_gp_poli,c02_bloat_parsimony_sean}, and elitism~\citep{c08_elitism_reduces_bloat_poli,a09_implicitly_controlling_bloat_gp_whigham}. All of these methods work directly on the genotype, with the result of modifying that genotype and intervening with the evolutionary process.

A recent review of simplification and bloat control techniques can be found in~\citep{a22_simplification_gp_javed}. Here, we focus on the former category of approaches, that are observing the runs, rather than intervening, because we believe that understanding the underlying phenomena is important to engineer better and faster evolutionary processes.

\subsection{Genotype and Phenotype - Definitions and Relations}
\label{ss_back_geno_pheno}

It is useful to recall a crisp definition of the genotype and of the phenotype in Genetic Programming. In this line of work, we discern them sharply by asking the following questions:
\begin{itemize}
    \item Which structure is manipulated by the genetic operators (mutation, crossover or other variation operators)? This we define as the genotype.
    \item Which structure determines behavior, i.e. has an impact on the output of the program/algorithm? This we define as the phenotype.
\end{itemize}

\begin{figure}[hb]
    \centering
    \includegraphics[width=\columnwidth]{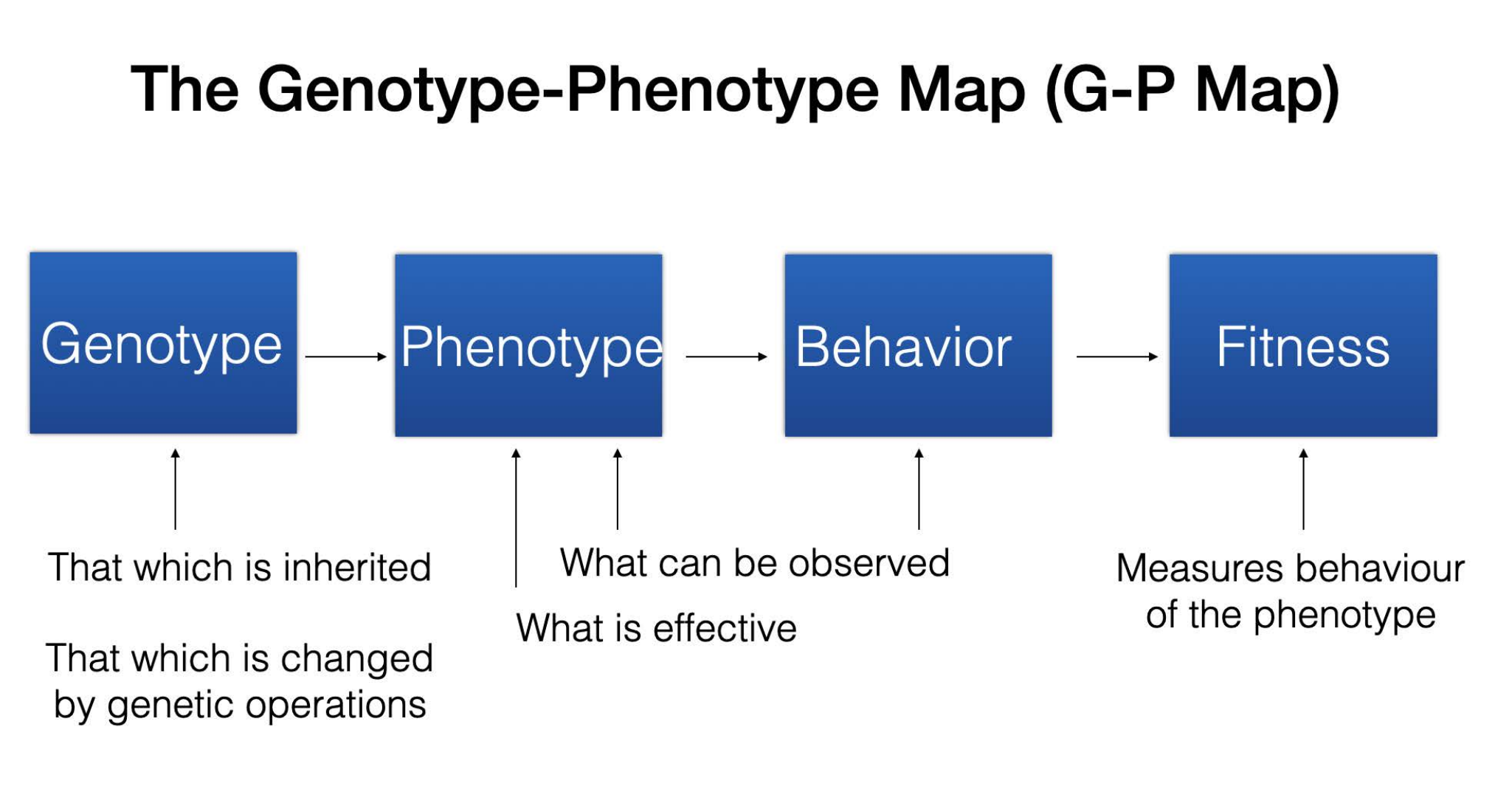}
    \caption{The relation between genotype, phenotype and behavior in Genetic Programming}
    \label{fig:geno-pheno}
\end{figure}

The situation is depicted in Figure~\ref{fig:geno-pheno}. The phenotype determines the behavior of an individual which in turn is judged by the fitness function.

In Biology, the notion of phenotype is somewhat vague. Anything that can be observed is considered the phenotype \citep{de2022phenotype}. So the definition of phenotype is relative to the intent of an observer. It could be one specific trait that is under observation, and thus the phenotype is restricted to that trait. Or it can be the entire organism. But in GP we can actually tie the phenotype to the part of the algorithm that makes a semantic contribution to the output. Thus we can tie it to the behavior of the algorithm that is then judged by the fitness function. 

Between genotype and phenotype is the so-called genotype-phenotype map (GPM). In GP, this is normally a relatively simple many-to-one mapping between genotypes and phenotypes. In Biology, however, the GPM can be much more complicated and typically is dynamic, a feature that few GP GPMs have (see \citep{cussat2019artificial} for examples). 
It has been argued that the GPM is a key ingredient in evolvability in living organisms~\citep{kirschner2005plausibility}. The GPM in GP, the mapping is normally many-to-one, because there are many more genotypes that
lead to the same structure with its subsequent behavior. 

It is easy to see the difference in a linear GP approach \citep{b07_lineargp_brameier_banzhaf}. As is well known, LGP individuals consist of sequences of instructions, some of which might not impact the overall behavior of an individual. In LGP this comes about because such non-effective instructions manipulate registers that have no impact on the output of a program. We can easily identify those instructions, and they do not belong to the phenotype of that individual. A similar, if less frequent phenomenon has been termed ''dormant nodes'' \citep{a10_dormancy_gp_jackson} in tree GP.
Only what is effective in terms of causing effective behavior is relevant and thus belongs to the phenotype. 

The problem for tree GP is that the parse trees we see under evolution are genotypes, not phenotypes, but somewhat hidden within such parse trees are the phenotypes of the TGP individuals. Thus, it is only a subset of nodes in a tree that effectively impacts its semantics or behavior, and it is that subset we are interested in when we want to understand what an individual evolving in TGP does. The problem of bloat in TGP is actually a problem of the growth of genotypes under evolution, not of the phenotypes. Provided we have a means to extract phenotypes from genotypes, we can hope to understand the solution semantics on the phenotype level.

The next section will describe an algorithm for extracting the phenotype of a TGP individual from its
genotype. We shall use this algorithm and its variants (which extract approximations to the phenotype) to argue that the relevant part of a GP individual, i.e., the part that requires explanation and understanding when solving a problem is actually only the phenotype and its behavior. Everything else is a distraction, perhaps necessary for the evolution of a solution, but not relevant for its behavior.

%% file: approach.tex
Most researchers consider the size of the tree representation as an approximate measure of the complexity of a candidate solution. Thus, a simplification procedure that can extract the phenotype from a genotype can reduce the complexity of a tree and enhance the potential interpretability of a solution while not changing the behavior, at least not significantly. In this study, the quality of a solution is measured as generalisation ability on real-world regression problems. For each tree under evolution (genotype) we shall derive its
phenotype and judge its behavior.

The impact of the simplification algorithm on the behavior of a solution will depend on the type of simplifications we apply to the genotype, and can be two-fold: (i) the behaviour is preserved, since the removed code has absolutely no impact on the behaviour, or (ii) the behaviour is changed, as the removed code contributed measurably to the behavior. 
In this study, we refer to the former as exact simplification, whereas the latter is referred to as approximate simplification. For approximate simplification, a problem-independent parameter $t$ is introduced to control for different levels of approximation
accuracy and represents the percentile of the distance distribution within a tree (see Section~\ref{ss_threshold} for more details).

Consider a tree $T$ implemented as a list of program elements. Given a set of training $n$ instances, the corresponding $n$-dimensional output vector is represented by $\hat y$ and is called its semantics. Let $T_s$ represent a given subtree in $T$, rooted at node $s$, and $\hat y_s$ the respective semantics. During fitness evaluation, calculating $\hat y$ requires calculating $\hat y_s$ for each $T_s$ in $T$. Specifically, the algorithm recursively traverses the tree in a bottom-up manner, starting from the deepest levels of $T$ and computes the outputs for each $T_s$ in $T$ up the tree. 

\begin{algorithm}[tbh]
	\begin{minipage}{\columnwidth}
			\begin{small}				
				\begin{enumerate}
\item Compute $\hat y_{s}\ \forall\ T_{s}\ \in\ T$. 
\item Given some similarity measure $f(x, y)$, create $M$.
\item Given some value of $t \geq 0$, reduce $M$ to semantically equivalent pairs (i.e., $\forall \ (\hat y_{s_i}, \hat y_{s_j}),\ f(\hat y_{s_i}, \hat y_{s_j}) \leq p(t)$).
\item While $M$ has entries:
\begin{enumerate}
    \item Select the largest tree $T_{s_i} \in M$.
    \item Select the smallest semantically equivalent tree $T_{s_j}$ for $T_{s_i}$.
    \item Replace $T_{s_i}$ with $T_{s_j}$. 
    \item Remove all trees contained by $T_{s_i}$ (i.e., $\forall\ T_{s_{i_j}} \in T_{s_i}$).
    \item Remove $T_{s_i}$ from $M$.                 
\end{enumerate}                        
\end{enumerate}
\end{small}
\end{minipage}
\caption{Pseudo-code for the proposed simplification method.}
\label{fig:pseudo_code_method}
\end{algorithm}

The proposed simplification algorithm capitalizes upon the mechanics of the fitness-evaluation algorithm. In this way, we can avoid redundant function calls and calculations. 
In particular, when $\hat y_s$ is calculated for each $T_s$, the semantics are temporarily stored in a dictionary $D$ of the form $D=\{idx_{s}:\ y_s\}$, where $idx_s$ represents the index of the node $s$ (the root of $T_s$). Note that any terminal node in $T$ should be also regarded as a subtree. 
We use $D$ to create a similarity matrix $M$ between every pair of semantics $y_{s_i}$ and $y_{s_j}$. 
In practice, we only calculate and store the lower triangular matrix since we assume that similarity $f(\hat y_{s_i}, \hat y_{s_j}) = f(\hat y_{s_j}, \hat y_{s_i})$ and $f(\hat y_{s_i}, \hat y_{s_i}) = 0$, where $f(x,\ y)$ is some similarity measure. 
Then, $M$ is filtered to store only semantically equivalent pairs. When the exact simplification is desired, $T_{s_j}$ is said to be equivalent to $T_{s_i}$ if $f(\hat y_{s_i}, \hat y_{s_j}) \cong 0$, where $\cong$ denotes equality within floating-point rounding errors. In the case of an approximation, the equivalence means $f(\hat y_{s_i}, \hat y_{s_j}) \leq p(t)$, $t>0$. We denote the exact equivalence (the exact phenotype) with $t=0$.
Additional meta-data is provided with $M$: the length of each subtree and whether one subtree contains another. 

The proposed algorithm will then use this information as follows: First, the largest tree $T_{s_i}$ is selected from $M$ and, for it, the smallest semantically equivalent tree $T_{s_j}$ is chosen. Second, $T_{s_i}$ is replaced by $T_{s_j}$. Finally, $T_{s_i}$ is removed from $M$ along with the subtrees it contained. The procedure is iterated while $M$ has entries.
Algorithm~\ref{fig:pseudo_code_method} describes the proposed simplification.

\subsection{Approximate simplification accuracy}
\label{ss_threshold}
We introduce a relaxation on the exact simplification condition which translates into an additional degree of simplification by coarse-graining the genotype. 
This is achieved via problem-independent parameter $t$ representing the $t^{th}$ percentile of the similarity distribution generated when pruning a given tree $T$. Specifically, when $M$ is constructed for a given tree, the values it contains represent a similarity distribution. In this sense, $p(t)$ is the value of the $t^{th}$ percentile of that distribution and inversely represents the approximation accuracy for the phenotype extraction. In our experiments, we explore four different values of $t$: 2.5\%, 5\%, 10\% and 20\%. 
Thus $T_{s_j}$ is said to be compatible to $T_{s_i}$ if $f(\hat y_{s_i}, \hat y_{s_j})\leq p(t)$.
Given the fact that the input data for different regression problems have different distributions, including the range of values of the target, by defining $p(t)$ as the $t^{th}$ percentile of the similarity distribution, a robust and problem-independent approach for approximate simplification is achieved. 
From this perspective, the exact simplification corresponds to $t=0$. To accommodate for the floating-point rounding error, in our experiments, two semantics vectors are considered \textit{equal} if the similarity between them, after being rounded to five decimals, equals zero.

\begin{figure}
    \centering
    \includegraphics[width=0.85\columnwidth]{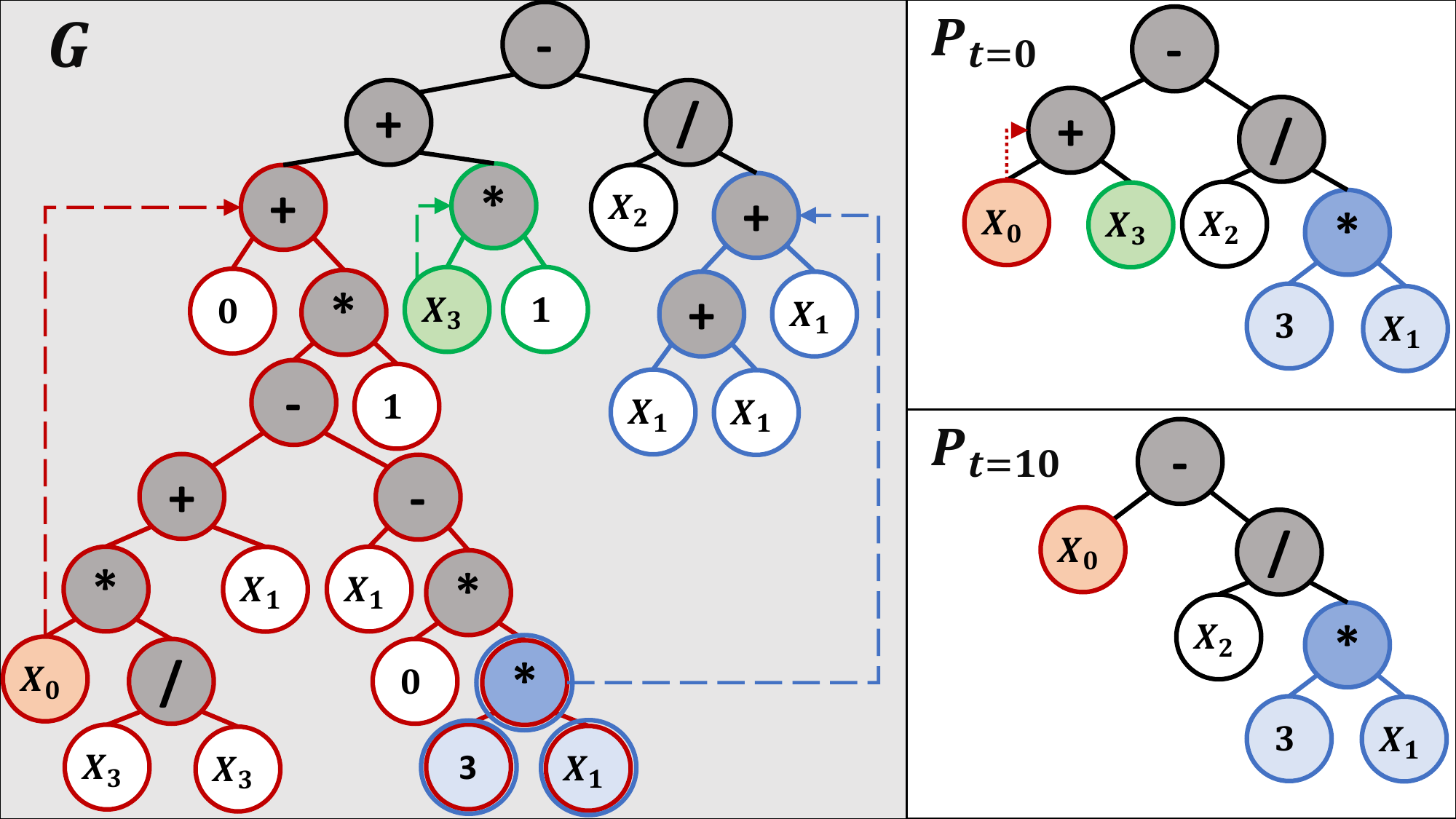}
    \caption{Illustrative mapping from genotype to phenotype. The genotype is represented with $G$ (left box), whereas the exact and approximate phenotypes are represented with $P_{t=0}$ and $P_{t=10}$ (right boxes).}
    \label{fig_pruning_example}
\end{figure}

Figure~\ref{fig_pruning_example} provides an example demonstrating the method's functioning. It is divided into three areas: $G$, $P_{t=0}$ and $P_{t=10}$. The $G$ area depicts the genotype, while the other two represent different phenotypes, simplifications depending on the value of $t$. For $t=0$, we perform exact simplification, thus extracting the true phenotype. For $t=10$, we perform approximate simplification.
From the figure, $t=0$ requires only three iterations of the algorithm (represented in red, blue and green subtrees, respectively). Note that, however, more redundancies are present within the subtrees. For example, at the bottom of the red subtree, on the right, we find the expression $0 \times 3 \times X_1 = 0$. However, $X_0$ is the smallest semantically compatible subtree of the big red subtree. Thus, the algorithm will first replace the red subtree with $X_0$, avoiding an unnecessary replacement of $0 \times 3 \times X_1$ with $0$.
Additionally, the algorithm can perform simplification by leveraging information from different branches. Look at the subtrees in blue. The semantics of the subtree $3 \times X_1$ is the same as that of $X_1 + X_1 + X_1$. Although both subtrees are on different branches, this method still identifies this redundancy and replaces the smallest with the largest. 

To understand what happens when $t=10$, let's assume that the result of adding $X_3$ to $X_0$ has a small effect on the overall behavior of $T$ because $X_3$ has \textit{small} values. Formally, $f(\hat y_{X_0}, \hat y_{+}) \leq p(10)$; whereas $f(\hat y_{X_3}, \hat y_{+}) > p(10)$. Thus, $X_0$ will replace the subtree rooted at $+$, resulting in a further simplification. Note that this time, however, we arrive at an approximation to the phenotype, smaller but semantically different from that obtained through exact simplification of the genotype. 

%% file: experiments.tex
\subsection{Datasets}
\label{ss_exp_data}
We assessed our method on five real-world regression problems, which were standardized to zero mean and a standard deviation of one. Major characteristics of these datasets can be found in Table~\ref{table: experimentDatasets}. 

\begin{table}[h]
\centering
\caption{Five datasets used in the empirical study.}
\label{table: experimentDatasets}
\begin{tabular}{|c | c | c | c | c|}
 \hline
 \textbf{Dataset} & \textbf{\#Instances} & \textbf{\#Features} & \textbf{Target range} \\ [0.5ex] 
 \hline\hline
 Boston~\citep{a78_housing_harrison} & 506 & 13 & [5, 50] \\
 \hline
 Bioav~\citep{c06_gp_bioav_archetti} & 358 & 241 & [0.4, 100.0] \\
 \hline
 Heating~\citep{a12_heating_load_tsanas} & 768 & 8 & [6.01, 43.1] \\
 \hline
 Slump~\citep{misc_concrete_slump_test_182} & 103 & 7 & [0, 29] \\
 \hline
 Strength~\citep{misc_concrete_compressive_strength_165} & 1005 & 8 & [0, 1145] \\
 \hline 
\end{tabular}
\end{table}

\subsection{Experimental settings}
\label{ss_exp_environment}
Table~\ref{tab_hyperparameters_list} lists the hyper-parameters (HPs) used in this study, along with cross-validation settings. The HPs were selected following the common practice found across the literature to avoid a computationally demanding tuning phase. 
No limit to tree depth was applied during evolution in order not to interfere with the evolutionary process and to allow bloat to reveal itself in all its \textit{splendour}. Although this resulted in a heavy computational demand during the experiments, it allowed us to perform an unbiased assessment of monitoring the phenotypes. 
We used two sets of experiments: (i) one with a low mutation rate and high crossover, and (ii) another with a high mutation rate and low crossover. 
A low mutation / high crossover encourages a more exploitative search, whereas a high mutation / low crossover tends towards exploration. These cases have different implications on the genotype of candidate solutions: High recombination of already existing genetic material in the former, and a higher disruption and novelty in the latter. Following previous studies about bloat~\citep{c95_complexity_compression_nordin,b96_explicitly_defined_introns_nordin}, we expected to observe more bloat to occur with higher crossover rates. 

We rely on GPOL~\citep{a21_gpol_bakurov} to conduct our experiments. GPOL is a flexible and efficient multi-purpose optimization library in Python that covers a wide range of stochastic iterative search algorithms, including GP. Its modular implementation allows for solving optimization problems, like the one in this study, and easily incorporates new methods. The library is open-source and can be found by following~\href{https://gitlab.com/ibakurov/general-purpose-optimization-library}{ \underline{this link}}. The implementation of the proposed approach can be found there.

\begin{table}[t]
\centering
\caption{Summary of the hyper-parameters. Note that $P(C)$ and $P(M)$ indicate the crossover and the mutation probabilities, respectively.}
\resizebox{.5\textwidth}{!}{
\begin{tabular}{|l|l|}
\hline
\textbf{Parameters}          & \textbf{Values}                                               \\ \hline\hline
\textnumero Train/test split & 70/30\%                                                       \\ \hline
\textnumero Cross-validation & Monte-Carlo                                                   \\ \hline
\textnumero runs             & 30                                                            \\ \hline
\textnumero~generations      & 100                                                           \\ \hline
Population's size            & 100                                                           \\ \hline
Functions ($F$)              & \{+, -, x, /\}                                                \\ \hline
Initialization               & RHH (max depth limit of 5)                                    \\ \hline
Selection                    & tournament with 4\% pressure                                  \\ \hline
Genetic operators            & \{swap crossover, subtree mutation\}                          \\ \hline
$P(C)$                       & \{0.8, 0.2\}                                                  \\ \hline
$P(M)$                       & \{0.2, 0.8\}                                                  \\ \hline
Maximum depth limit          & Not applied                                                   \\ \hline
Stopping criteria            & Maximum \textnumero~generations                                       \\ \hline
\end{tabular}}
\label{tab_hyperparameters_list}
\end{table}

\subsection{Experimental findings}
\label{ss_exp_findings}
This section demonstrates the effects of the GPM on the population dynamics across five perspectives: size, diversity, fitness, deviation between the genotype and phenotype semantics, and variation of the proportion of terminals. Each perspective is studied with two different mutation/crossover rates (low/high and high/low), provided as sub-figures.  
The figures report different perspectives aggregated across all the problems, allowing therefore for a problem-independent analysis and interpretation.
The genotype is represented by standard TGP, and it is used as a baseline. In the plots, it is represented as a gray line. The exact phenotype is represented with a black dashed line, whereas the approximate phenotypes are shown in green, red, blue and black solid lines, referring to different degrees of approximation (2.5\%, 5\%, 10\%, 20\%, respectively).

\subsubsection{Size}
\label{ss_exp_findings_s1}

\begin{figure}
    \centering
    \includegraphics[width=\columnwidth]{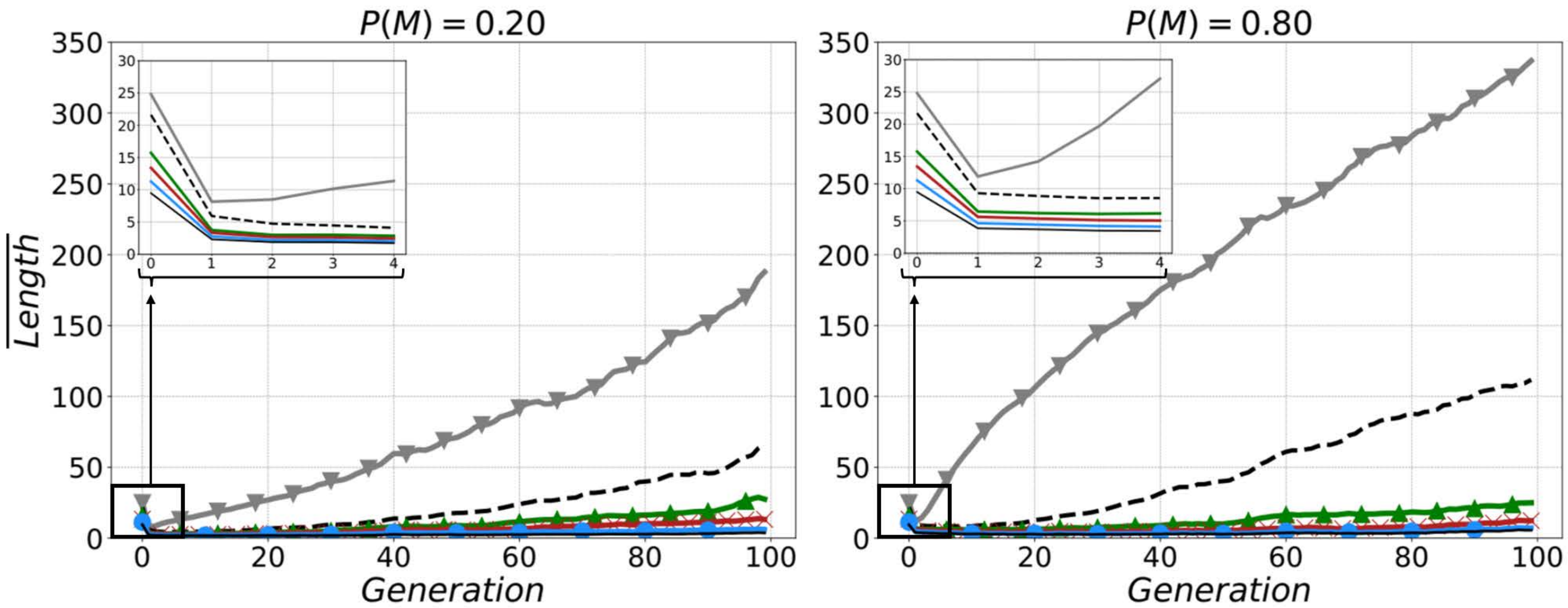}
    \caption{Growing average population length of genotypes, exact and approximate phenotypes, except for first generations (inset). Left: low mutation/high crossover; Right: high mutation/low crossover. 
    }
    \label{fig_len_pop_avg}
\end{figure}

Figure~\ref{fig_len_pop_avg} shows the average population length over 100 generations. The subplot on the left shows the low mutation/high crossover HP setting. On the right, high mutation/low crossover setting can be found. The inception plots provide a closer look at the first 5 generations. Analysis of the figure suggests that:
\begin{itemize}
    \item Trees in S-TGP (here, the genotype) grow quickly, except for the first 1-2 generations.
    \item Phenotypes also grow after some generations, but much slower, with the approximate phenotypes ordered according to the approximation accuracy $t$. 
    \item  In early generations, however, average length of the individuals decreases quickly then slower for phenotypes, but genotypes (what is observed in standard GP) only from the random start population to the first generation.
    \item The average length of the individuals is larger when a higher mutation rate is used for both genotypes and phenotypes. 
    \item The latter observation suggests to us that mutation, unlike what was expected, is more destructive than crossover for the applications considered. This makes the evolutionary process foster bloat as a protective measure. Similar results were found by~\citep{a02_causes_of_growth_soule} where higher mutation provoked larger growth.    
    \item Monitoring the exact phenotype shows a reduction in average length of individuals by a factor of 2-3, compared to the genotype. 
    \item Approximate phenotypes show a 10 to 20-fold reduction in average length of individuals. 
    \item The sudden drop in the size of individuals in the first generation is an indication that selection prefers small and fitter individuals out of the randomly composed trees by Ramped-Half-and-Half. Phenotypes stay close or drop even further in subsequent generations.
\end{itemize}

\subsubsection{Semantics of Approximate Phenotypes}
\label{ss_exp_findings_s2}
\begin{figure}
    \centering
    \includegraphics[width=\columnwidth]{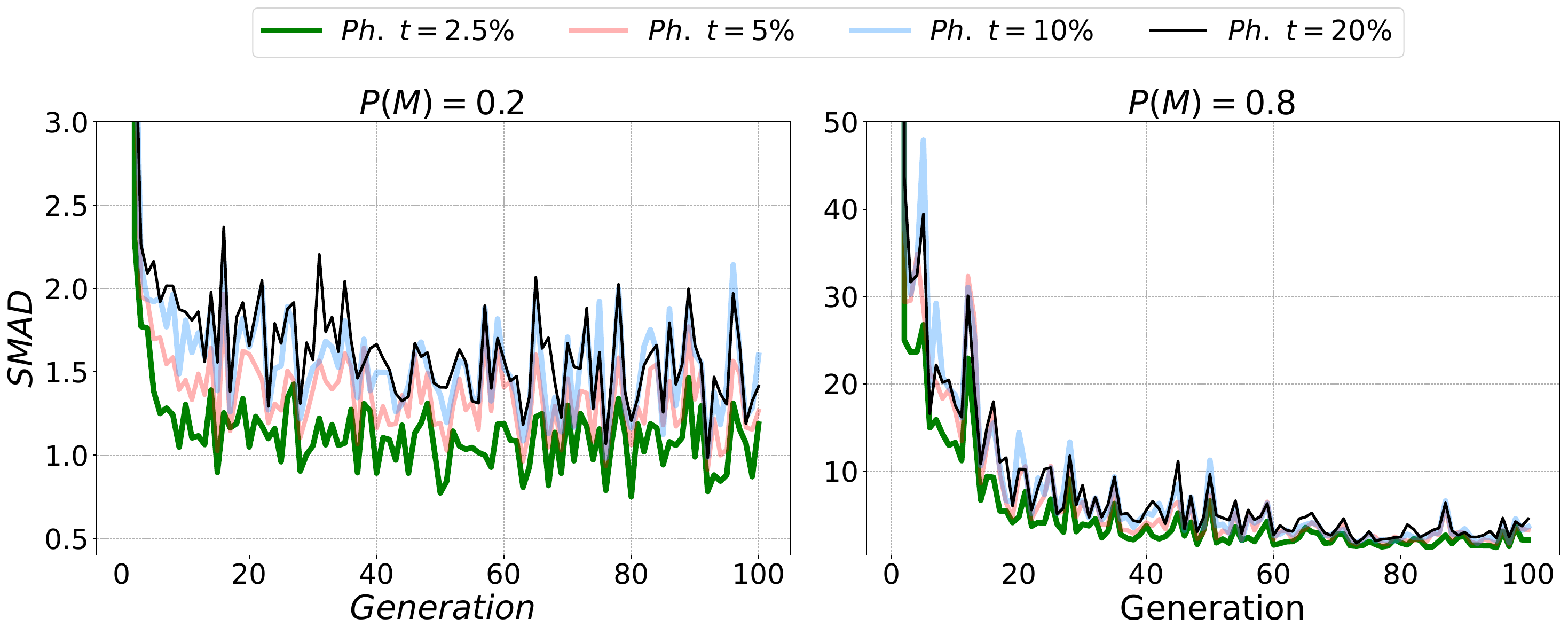}
    \caption{Mean absolute deviation between the semantics of the genotype and the approximated phenotypes. 
    Note the difference in scale.
    }
    \label{fig_smad_pop_median}
\end{figure}

Figure~\ref{fig_smad_pop_median} depicts the semantics mean absolute deviation (SMAD) between the genotype ($T$) and the respective phenotype approximations ($T_{t}, \ t \in \{2.5,\ 5,\ 10,\ 20\}$). Specifically, given the training dataset $X$ with $n$ training instances, and the respective outputs of $T$ and $T_{t}$, represented as $\hat{y}_{T}$ and $\hat{y}_{T_{t}}$, $MAD(\hat{y_T},\ \hat{y_{T_{t}}}) = \frac{ \sum |\hat y_{T} - \hat y_{T_{t}}| } {n}$.
To enhance the patterns, 
we neglect the line markers and apply a higher degree of transparency to those lines representing the intermediate approximations ($t=5\%$ and $t=10\%$). 
On the left, the low mutation/high crossover HP setting is shown, whereas the high mutation/low crossover setting can be found on the right.
From the figure, one can notice that:
\begin{itemize}
    \item In all cases, there is a notable decrease in SMAD in the first generations. This can be associated with the fact early in evolution that trees are still more random combinations of program elements rather than refined input-output mappings. 
    Thus, even mild modification of their structure results in substantial semantic perturbations when compared to those in later stages of the evolution. 
    \item In general terms, SMAD exhibits an exponential decay. For $P(M)=0.2$, it is more pronounced whereas for $P(M)=0.8$ the decay is slower.
    \item For both mutation rates, one can conclude that the larger the degree of phenotypic approximation, the larger the SMAD.
    \item The difference between $t=2.5\%$ and $t=20\%$ is particularly notable: a higher degree of approximation results in bolder simplification and, thus larger SMAD.
    \item When $P(M)=0.8$, the SMAD tends to exhibit significantly larger values, see the scale of subfigure is different by an order of magnitude.
    \item This fact might unveil the nature of redundancies between different HPs. Recall that when $P(M)=0.2$, one expects less exploration diversity, while with $P(M)=0.8$, a larger diversity is expected. 
    \item More disruptive changes in behavior happen when the search is more explorative.
\end{itemize} 

\subsubsection{Diversity}
\label{ss_exp_findings_s3}
\begin{figure}
    \centering
    \includegraphics[width=\columnwidth]{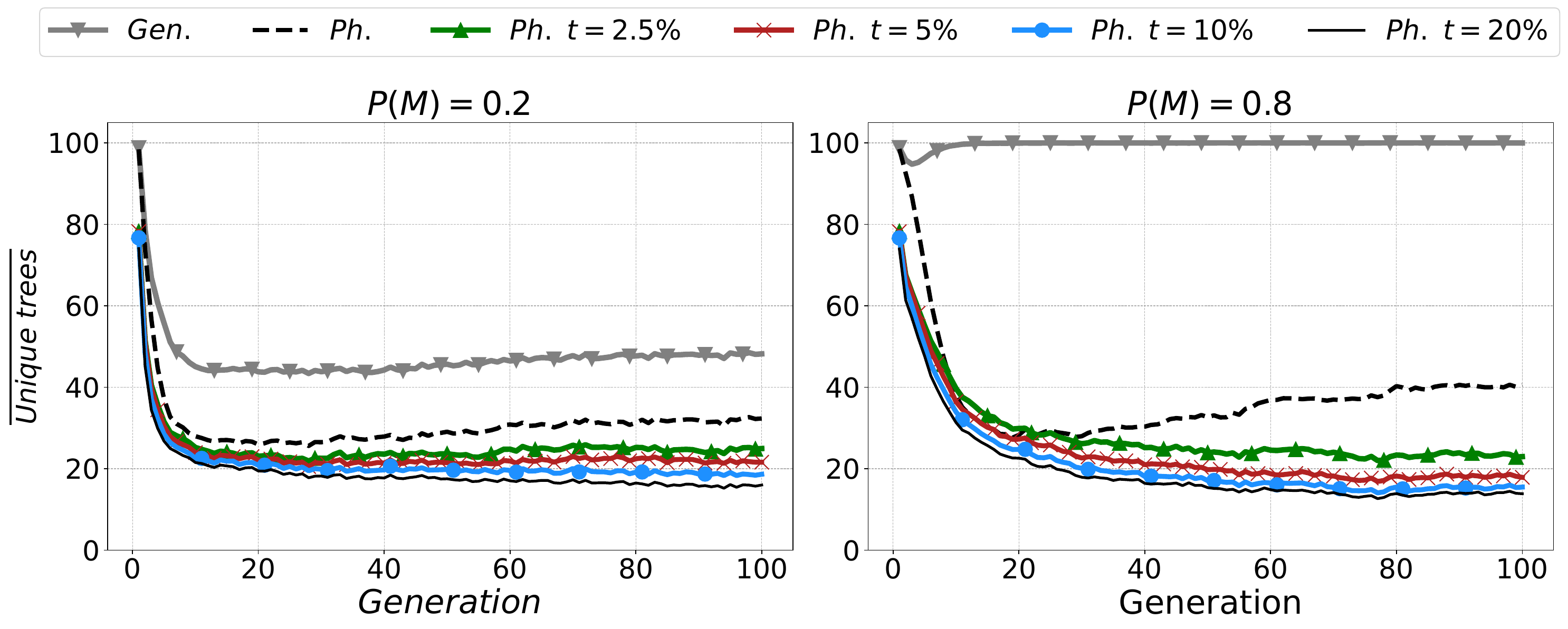}
    \caption{Diversity across evolutionary runs. Genotyic diversity stays high for high mutation scenario.}%
    \label{fig_diversity_pop_median}
\end{figure}

Population diversity has been long considered an important characteristic of the evolutionary process. Losing diversity, especially at the beginning of the evolution, frequently translates into premature convergence. We here examine the genotypic diversity and it is defined as the number of unique trees per generation.
Figure~\ref{fig_diversity_pop_median} depicts the mean of genotypic diversity with columns representing the two HP settings: low mutation/high crossover on the left, and high mutation/low crossover on the right.

Analysis of the figure suggests that:
\begin{itemize}
    \item During the first few generations we find a notable decrease in genotypic diversity. The decrease is more dramatic, however, for $P(M)=0.2$ as less exploration takes place.
    \item Different lines seem to stack one upon the other and the order is a function of the approximation degree. In other words, the larger the phenotype approximation, the smaller the diversity of the resulting population.   
    \item In general terms, the diversity exhibits an exponential decay. For a higher mutation setting, the decay is slower as more exploration takes place. 
    \item There is an exception to the trend lines, however, when looking at the genotypic diversity for $P(M)=0.8$. There, the individual genotypes tend to be distinct in terms of their genomes across the evolutionary run. 
    \item After 40-50 generations of the evolutionary process, the number of unique trees for the approximated phenotypes converges to a small range, around 15-25. This is observed for both for $P(M)=0.2$ and $P(M)=0.8$. 
    In other words, the genotypic diversity of the populations consisting of the approximated phenotype converges to the same level, regardless of the crossover/mutation rates. 
    \item We can observe a growing trend for the exact phenotype (black dashed line) after 20 generations. This can be an indication that even the phenotype is not protected against bloat. Note that this phenomenon is more pronounced in the high-mutation setup, which was found to be associated with higher bloat in Figure~\ref{fig_len_pop_avg} (possibly caused by destructive effects of the mutation operator).
\end{itemize} 

\subsubsection{Fitness}
\label{ss_exp_findings_s4}
\begin{figure}
    \centering
    \includegraphics[width=\columnwidth]{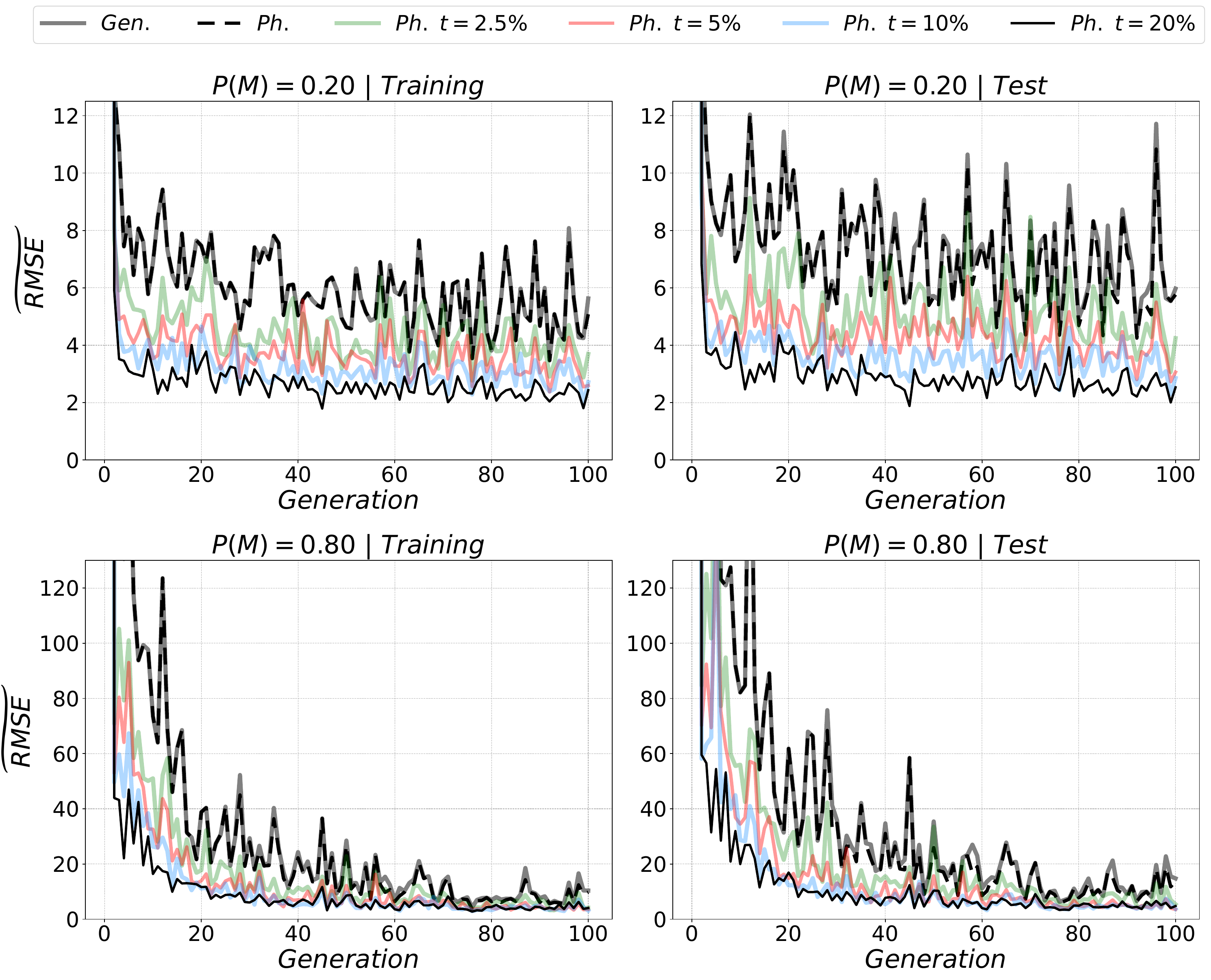}
    \caption{Median fitness in the population. Approximate phenotypes exhibit better fitness.}
    \label{fig_fit_pop_med}
\end{figure}

For each generation, we record the average population fitness, for both training and test data. Fitness here is error, so smaller is better. Figure~\ref{fig_fit_pop_med} depicts median values for the two different mutation/crossover rates, represented here in distinct rows. Different columns represent distinct data partitions: training data on the left, and test on the right. 
The line markers were removed and a higher line transparency was used for intermediate levels of phenotype approximations to highlight the trend. 
The analysis of the figure suggests that:
\begin{itemize}
    \item There is a dramatic improvement in population fitness in the first few generations, which can be related to diversity loss.
    \item In general terms, fitness/error exhibits an exponential reduction. For a higher mutation setting, the decay is slower as more exploration takes place.      
    \item The smaller the phenotype approximation accuracy, the better the fitness of the resulting individuals. That is to say, approximated phenotypes in the population tend to exhibit better fitness. 
    \item Fitness progress achieved with $P(M)=0.8$ are notably larger than those observed for $P(M)=0.2$, particularly in the first half of runs. 
    \item The conclusions hold for both training and test data.    
\end{itemize}

\begin{figure}
    \centering
      \includegraphics[width=\columnwidth]{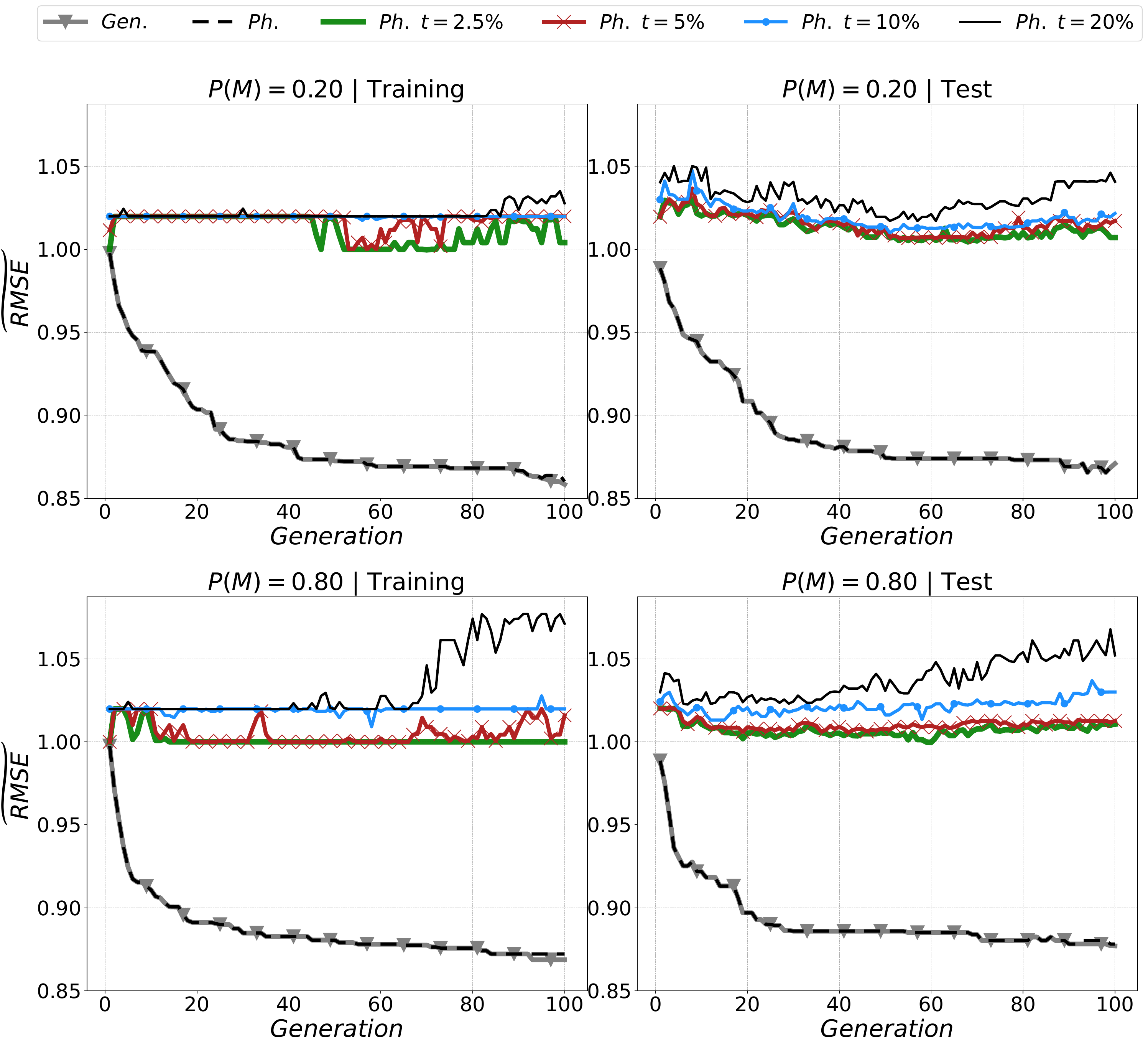}
    \caption{Median fitness of elite individuals. The fitness of phenotypes deteriorates with approximation.}
    \label{fig_fit_elite_median}
\end{figure}

Figure~\ref{fig_fit_elite_median} shows the median fitness of the best/elite individuals (selected on the training data). The structure of the figure conforms to that of Figure~\ref{fig_fit_pop_med}. Unlike what was observed with population fitness, larger phenotypic approximation results in virtually no progress in elite fitness.

It is important to clarify the nature of the overlap between the lines representing the exact phenotype (black dashed line) and the genotype (gray solid line): Their semantics are expected to be the same. There is, however, some rare mismatch as a consequence of minor floating-point rounding errors that occasionally occur in large trees,  percolating through the tree. From all the generated results, these correspond to 1\% of cases and mostly regard one of the problems.

\subsubsection{Structure}
\label{ss_exp_findings_s5}
\begin{figure}
    \centering
      \includegraphics[width=\columnwidth]{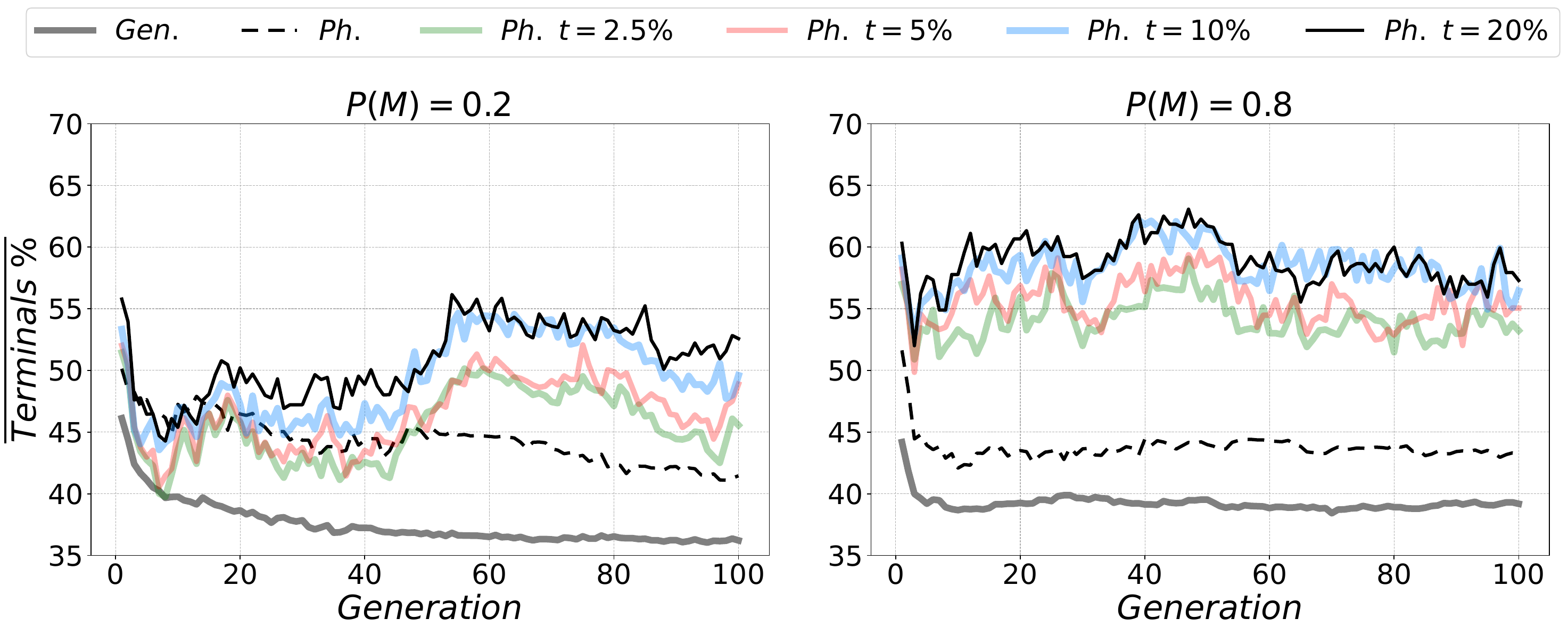}
    \caption{Terminal nodes percentage in the elite trees. Being smaller, phenotypes have a higher percentage of terminals.}
    \label{fig_program_elements_pop_average}
\end{figure}

Figure~\ref{fig_program_elements_pop_average} shows the proportion of terminals (input features and constants) in trees during evolutionary runs. The rest of the nodes are operators. We can observe that:
\begin{itemize}
    \item Genotypes exhibit a notable reduction of terminal proportion in the first few generations. This is associated with a growth in tree depth, reflected in a larger amount of function nodes.
    \item Later in the search, the proportion of terminals stabilizes around 35-40\%.
    \item As was previously observed in other figures, lines tend to follow an order as a function of the approximation degree. 
    \item The rougher the approximation, the larger the proportion of terminals to functions. This is expected since rougher approximation results in smaller phenotypic trees, with a smaller number of levels and fewer function nodes.
    \item A larger proportion of terminals in the phenotypes can be observed with a higher mutation rate (suplot on the right). We consider this to be related to the fact that the extent of simplification is larger for this setting, as it was already observed in Figure~\ref{fig_len_pop_avg}.
\end{itemize}

\begin{figure}
    \centering
      \includegraphics[width=0.85\columnwidth]{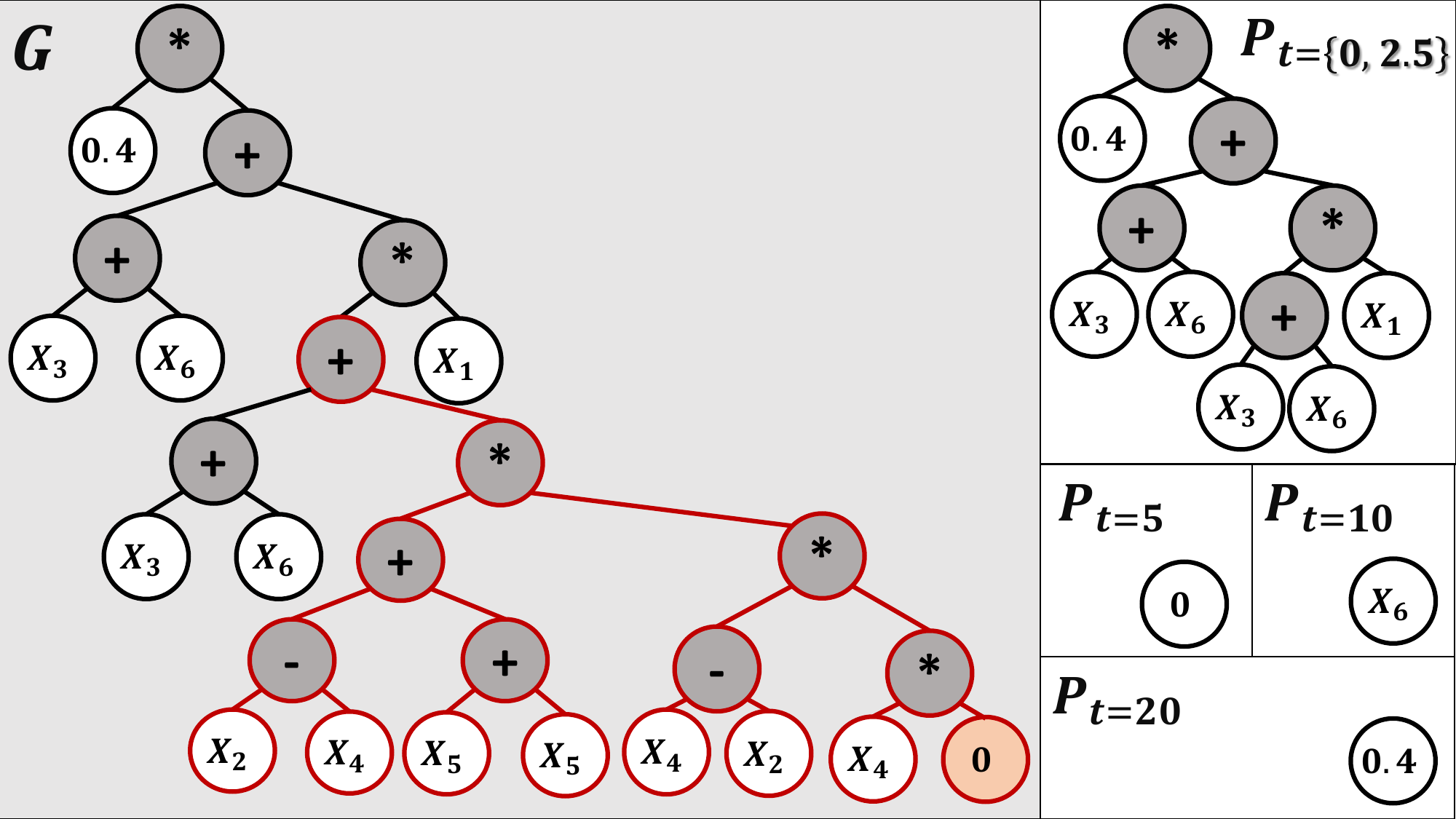}
    \caption{Genotype and exact/approximated phenotypes from a run on the Slump problem. See text for details.}
    \label{fig_real_phenotype}
\end{figure}

Figure~\ref{fig_real_phenotype} provides an example of the exact ($t=0$) and approximated ($t=\{2.5, 5, 10, 20\}$) phenotype(s) for the elite observed at generation 95 of run 2 for the Slump problem. The genotype ($G$) is represented in the gray box. The exact phenotype is represented in the upper left box ($P_{\{0,\ 2.5\}}$), obtained by removing the red subtree, due to zero constant multiplication (nodes circled red). This happens to be the same phenotype as for the approximate phenotype with $t=2.5\%$ which did not produce a smaller tree. The rougher approximations consist of single-node trees, which are of small utility for this application. 

%% file: conclusions.tex

%
%
%
 

Traditionally, only the genotype has been considered
in Tree Genetic Programming (TGP). Here we study the notion of phenotype and its relationship to the genotype in TGP to clarify what evolved solutions truly predict. We developed a unique simplification method that maps a genotype to its phenotype by removing semantically ineffective code from the genotype. This mapping shows that phenotypes are hidden within the larger genotypic trees but can be extracted via simplification, achieving a size reduction of up to 20 folds, which facilitates understanding. 
We also study how simplification affects population dynamics.  
Particularly, analysis of the diversity suggests that the genotype population is normally based on a small number of unique phenotypes, even when evolution is more explorative. 
Moreover, when the approximated phenotype is extracted, larger semantic perturbations are observed in the early stages of evolution.
Finally, approximated phenotypes allow for better population fitness, although elite solutions show deterioration when approximated. While this constitutes valuable insights for future research directions, such as the development of new bloat control methods, we are now working on improving the efficiency of the simplification algorithm.